\RequirePackage[T1]{fontenc}

\documentclass[journal]{IEEEtran}
\usepackage{url}
\usepackage{booktabs}
\usepackage{amsmath}
\usepackage{amssymb}
\usepackage{amsfonts}
\usepackage{graphicx}
\usepackage{microtype}
\usepackage{xcolor}
\usepackage{colortbl}
\usepackage{tabularx}
\usepackage{array}
\usepackage{multirow}
\usepackage{placeins}
\usepackage{pifont}
\usepackage{cite}
\usepackage{etoolbox}
\usepackage{flushend}
\usepackage{hyperref}
\hypersetup{hidelinks}

\title{MT-WAM: Reorienting the One-Pass Predictive Representation Toward Action Generation}

\author{Yiguang Yang\textsuperscript{1,2}, Jiankun Peng\textsuperscript{1,2},
Xiaoming Wang\textsuperscript{1,2}, Yiran Zhang\textsuperscript{1,2},
and Zhibo Fang\textsuperscript{1}\\[0.5em]
\small\textsuperscript{1}Aerospace Information Research Institute, Chinese Academy of Sciences\\
\small\textsuperscript{2}University of Chinese Academy of Sciences}

\IEEEaftertitletext{\vspace{-1.5\baselineskip}}

\newcommand{\method}{\mbox{\textsc{MT-WAM}}}
\newcommand{\fastwam}{\textsc{Fast-WAM}}
\newcommand{\liberoplus}{\textsc{LIBERO-Plus}}
\newcommand{\robotwin}{\textsc{RoboTwin 2.0}}
\newcommand{\best}[1]{\textbf{#1}}
\newcommand{\second}[1]{\underline{#1}}
\newcommand{\cmark}{\ding{51}}
\newcommand{\xmark}{\ding{55}}
\definecolor{tablehighlight}{gray}{0.92}
\newcommand{\tablehighlightsetup}{%
  \setlength{\aboverulesep}{0pt}%
  \setlength{\belowrulesep}{0pt}%
  \setlength{\extrarowheight}{1.2pt}%
}

\begin{document}

\bstctlcite{IEEEexample:BSTcontrol}

\maketitle

\begin{abstract}
\fastwam{} shows that video--action co-training improves control without generating future video at inference, making the representation from a single video diffusion Transformer forward central to action generation. However, future-observation prediction does not explicitly prioritize the future dynamics and visual structure needed for control. We present \method{}, which retains the original training objectives and adds complementary supervision for future two-dimensional point trajectories and visual features. A lightweight dual-stream branch copied from the video backbone's final blocks provides target-specific processing, while a structured attention mask prevents cross-stream attention. Motion-stream tokens supply additional dynamics conditions to the action expert. Future visual-feature prediction provides supervision in a feature space that captures object and spatial structure. This supervision trains the video backbone to provide more informative visual context for action generation under changing visual conditions, without adding visual-feature-stream tokens to action conditioning. At inference, \method{} uses video and motion caches computed once per replan and skips future-video prediction. Without additional embodied policy pretraining, \method{} achieves 98.2\% success on LIBERO and 73.7\% on \liberoplus{}, exceeding \fastwam{} by 23.8 percentage points on the latter. On \robotwin{} Clean2Rand, Random success increases from 6.30\% to 19.40\%; across four real-world tasks, average success increases from 67.0\% to 77.8\%.
\mbox{Code repository: \url{https://github.com/Alexi1984/MT-WAM}.}
\end{abstract}


\section{Introduction}

\IEEEPARstart{R}{obot} manipulation requires policies to anticipate how a scene will change during interaction and use that anticipation to generate actions. World action models (WAMs) couple future-observation prediction with action generation, allowing visual and dynamic priors acquired through video pretraining to shape representations for control~\cite{yuan2026fastwam}. These priors help policies model scene evolution, object interactions, and state transitions, and support generalization across tasks and visual conditions. Many WAMs instantiate and denoise future video tokens at inference, using the representations formed during explicit future modeling to condition action generation. \fastwam{} separates the contribution of training-time video prediction from that of inference-time future generation through controlled comparisons~\cite{yuan2026fastwam}. It shows that explicit future generation can be omitted while retaining the control benefits of video--action co-training. The video diffusion Transformer (Video DiT) processes the current observation once and populates its per-layer key--value (KV) caches, which the action expert reuses throughout iterative action denoising. \fastwam{} achieves performance comparable to explicit future-imagination variants while substantially reducing inference latency. At the same time, its controlled comparisons show that removing video--action co-training causes a large performance drop. Once future generation is skipped, all predictive information from the video model reaches the action expert through this one-pass representation. The representation therefore becomes the critical interface between future-observation prediction and action generation.

The central question is whether this interface makes the information needed for action generation sufficiently prominent. Generating an action requires identifying the relevant objects, anticipating how they will move and interact, and estimating the resulting state transitions. The original future-observation prediction objective is not designed specifically for this purpose. Reconstructing a complete future observation requires the Video DiT hidden states to encode future motion and interaction progress together with the colors, textures, illumination, backgrounds, and other visual details needed to reproduce the future scene. Some of these visual details are not directly relevant to the next action, yet they still shape the same hidden states. The original objective therefore does not prioritize control-relevant predictive information in the representation supplied to the action expert~\cite{lambert2020objective,fu2021tia,nilaksh2026reconstruction,yu2026decision}.

\begin{figure*}[t]
  \centering
  \includegraphics[width=0.91\textwidth]{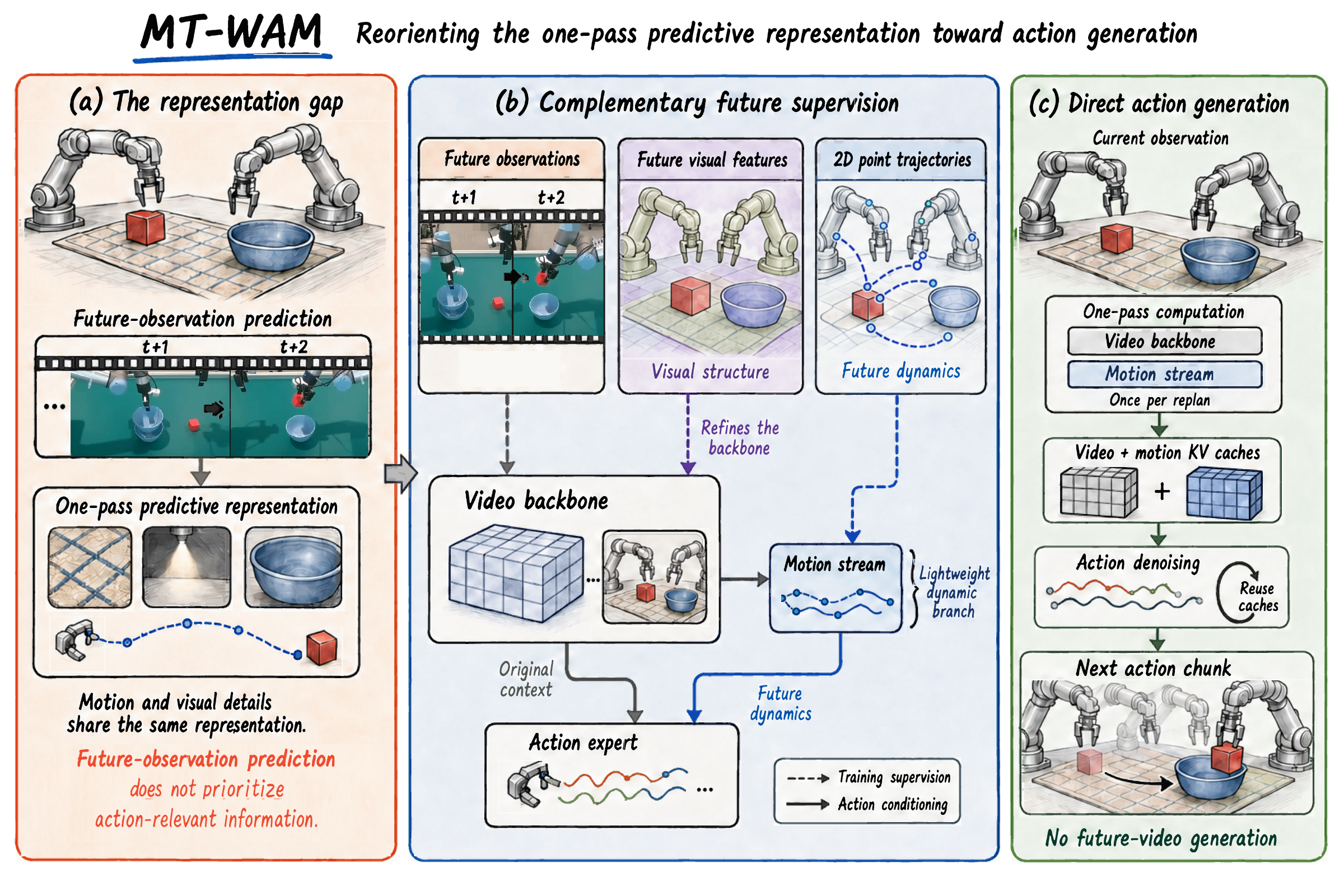}
  \caption{\textbf{MT-WAM overview.} \method{} augments future-observation prediction with 2D point-trajectory and future visual-feature targets through a dual-stream dynamic branch. Motion-stream tokens supply additional dynamics conditions, while future visual-feature supervision shapes the scene context that the video backbone provides to the action expert.}
  \label{fig:teaser}
\end{figure*}

Building on this view, we present \method{}, which improves the one-pass predictive representation through complementary supervision for future dynamics and visual structure (Fig.~\ref{fig:teaser}). It retains video--action co-training and the original future-observation prediction objective while adding future 2D point-trajectory and visual-feature prediction. CoTracker3~\cite{karaev2024cotracker3} provides 2D point-trajectory targets that describe image-space motion and supervise object motion, interaction progress, and state transitions relevant to action generation. Future visual-feature targets extracted by a frozen DINOv2 encoder~\cite{oquab2024dinov2} complement future-observation prediction with semantic information that is relatively robust to low-level appearance variation.

\begin{figure*}[!t]
  \centering
  \includegraphics[width=\textwidth]{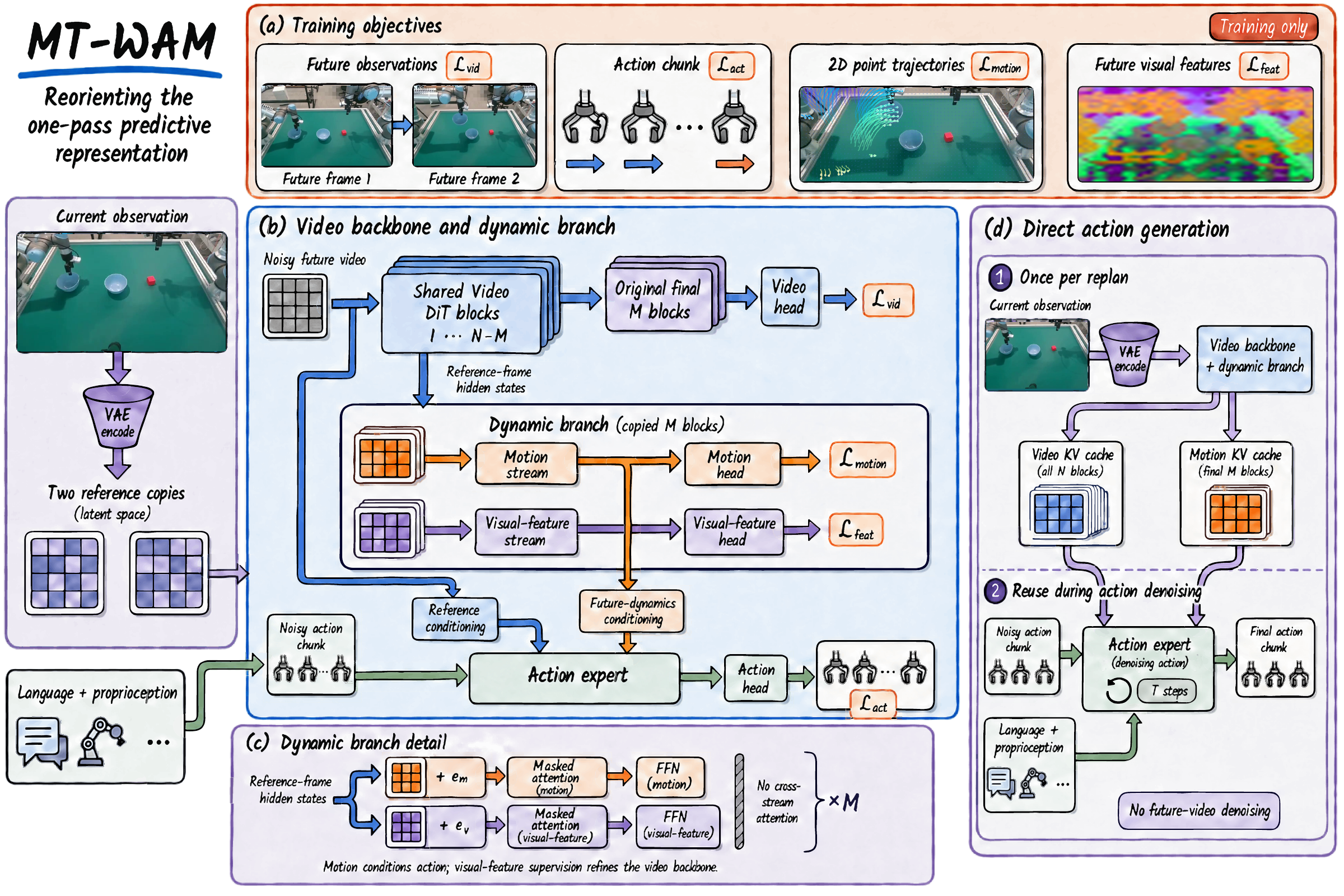}
  \caption{\textbf{MT-WAM training, architecture, and inference.} (a) Training uses future-observation, action, 2D point-trajectory, and future visual-feature objectives. (b) The video path retains future-observation prediction, the action expert generates actions, and an $M$-block dynamic branch predicts the two added targets. (c) The streams use separate FFNs and a structured attention mask; action queries attend to motion-stream tokens but not visual-feature-stream tokens. (d) The video backbone and dynamic branch populate the video and motion KV caches once per replan, and the action expert reuses both throughout denoising. The future-video, motion, and visual-feature heads are training-only; Fig.~\ref{fig:masks} shows the complete masks.}
  \label{fig:architecture}
\end{figure*}

Separate prediction heads leave the preceding Video DiT transformations shared across these heterogeneous targets, whereas a complete Transformer expert for each target duplicates the entire processing path, increasing parameter count and training cost. \method{} therefore copies the final $M$ Video DiT blocks to form a dynamic branch (Fig.~\ref{fig:architecture}), while the preceding blocks remain shared across the three prediction objectives. Two copies of the reference-frame latents distribute the demands of future-observation prediction, action generation, and the two additional objectives. Within the branch, the motion and visual-feature streams are initialized from the reference-frame hidden states available to the branch and use different learnable stream embeddings and independently updated stream-specific feed-forward networks (FFNs). A structured attention mask prevents the streams from attending to each other. Action queries attend to the motion-stream tokens, adding future-dynamics information to the action expert's original conditioning context, but do not attend to the visual-feature-stream tokens. Future visual-feature prediction encourages the Video DiT to capture semantic cues for identifying objects, locating interaction regions, and anticipating scene changes. Learning to predict future scene features thus trains the backbone to provide informative visual context for action generation under varying appearance and scene configurations. The motion and visual-feature prediction heads are used only during training. At inference, \method{} skips future video prediction and directly generates actions using the video and motion KV caches computed once per replan.

We evaluated \method{} on LIBERO, \liberoplus{}, \robotwin{}, and four real-world single- and dual-arm manipulation tasks. Without additional embodied policy pretraining, \method{} increased the \liberoplus{} Overall success rate by 23.8 percentage points, the \robotwin{} Clean2Rand Random success rate by 13.10 points, and the average real-world success rate by 10.8 points relative to \fastwam{}. The gains are more pronounced under distribution shift and in real-world manipulation than on the standard simulation benchmarks.

\begingroup
\setlength{\parskip}{0pt}

Our main contributions are threefold:
\begin{itemize}[\setlength{\topsep}{0pt}\setlength{\partopsep}{0pt}\setlength{\parsep}{0pt}]
  \item \textbf{Action-oriented learning of one-pass predictive representations.} We introduce \method{}, which improves the one-pass predictive representation through a joint design of complementary future prediction targets, target-specific processing, and distinct connections to action learning.
  \item \textbf{Lightweight target-specific processing and selective action conditioning.} \method{} uses a partially copied dynamic branch, stream-specific feed-forward networks, and a structured attention mask to process heterogeneous future targets. Future visual-feature prediction shapes the visual context produced by the backbone, complementing the dynamics conditions supplied directly by the motion stream. Together, these pathways connect predictive learning to both the visual and dynamic information used for action generation. \method{} retains the direct action-generation interface at inference.
  \item \textbf{Evaluation across simulation, out-of-distribution (OOD) settings, and real-world manipulation.} Without additional embodied policy pretraining, \method{} achieves a 98.2\% success rate on LIBERO and improves over \fastwam{} by 23.8, 13.10, and 10.8 percentage points on \liberoplus{}, the \robotwin{} Clean2Rand Random condition, and four real-world tasks, respectively. Controlled ablations support the asymmetric roles of the two branch streams.
\end{itemize}
\endgroup

\section{Related Work}

\subsection{Vision-Language-Action Policies}
Vision-language-action (VLA) policies map visual observations, language instructions, and robot state to actions by combining pretrained backbones with robot demonstrations~\cite{brohan2022rt1,brohan2023rt2,driess2023palme,oxe2023,octo2024,kim2024openvla}. Action chunking and diffusion- or flow-based heads support direct action generation~\cite{zhao2023act,chi2023diffusionpolicy,black2024pi0}. \method{} retains this interface but trains its video backbone with future-observation prediction and two complementary predictive targets.

\subsection{World Action Models}
World action models (WAMs) couple future-observation prediction with action generation so that video priors shape representations for control~\cite{yuan2026fastwam}. Many explicitly generate future visual trajectories or jointly denoise future video and actions. \fastwam{} retains video prediction during training but generates actions from the video KV cache produced by a single backbone forward~\cite{yuan2026fastwam}. ImageWAM similarly conditions action generation on KV caches produced by an image-editing backbone~\cite{zhang2026imagewam}. \method{} retains the one-pass interface of \fastwam{} and focuses on improving the representation produced by that pass.

\subsection{Beyond RGB Predictive Supervision in VLA and World Models}
VLA policies use auxiliary prediction and feature distillation to inform action generation. DreamVLA predicts future dynamic-region content, depth, and semantics with separate queries~\cite{zhang2025dreamvla}. FoMoVLA trains foresight queries through future-feature prediction and uses their hidden states to condition 2D point-trajectory prediction and action generation~\cite{li2026fomovla}. Track4Action distills world-centric 3D tracker features from action-aligned clips into queries used by its action head~\cite{wang2026track4action}.

In WAMs, AGRA aligns video hidden states used for action conditioning with pretrained visual features~\cite{qiu2026agra}, GeoSem-WAM adds dense prediction heads for future geometry and semantic segmentation~\cite{ma2026geosem}, and EgoWAM compares alternative pixel, visual-feature, and camera-stabilized 3D-motion targets for human--robot co-training~\cite{li2026egowam}.

JOPAT jointly denoises future visual latents, 2D point tracks, and actions while predicting track visibility~\cite{guan2026jopat}, whereas FlowWAM generates optical-flow videos as an action representation~\cite{chen2026flowwam}. For target-specific processing, X-WAM copies Video DiT tail blocks for depth prediction~\cite{guo2026xwam}, and MECo-WAM adds a separate training-time 4D expert with temporary current-frame geometry conditioning~\cite{zhang2026meco}. DreamWAM jointly denoises RGB and optical-flow latents, while geometry and semantic supervision refine video hidden states through gated residual branches~\cite{yuan2026dreamwam}.

Together, these studies highlight target-specific processing and action conditioning alongside the choice of predictive targets. \method{} combines a partially copied Video DiT tail with asymmetric action conditioning while skipping future prediction at inference. Motion-stream tokens supplement the action expert's original conditioning inputs, whereas visual-feature supervision refines the shared video backbone without adding visual-feature-stream tokens to action conditioning.

\section{Method}

\subsection{Problem Formulation}
\label{sec:problem-formulation}

We consider a WAM with a direct-policy interface~\cite{yuan2026fastwam}. At each decision step, the policy receives a current multi-view RGB observation $I_0$, a proprioceptive state $q_0$, and a language instruction $l$, and predicts an action chunk $a_{1:H}$ of horizon $H$. At inference, the video backbone processes $I_0$ once to compute the video key--value (KV) cache. Together with $q_0$ and $l$, this cache forms the action expert's original conditioning inputs, denoted by $\mathcal{C}_{\mathrm{base}}$:
\begin{equation}
  p_\theta(a_{1:H}\mid I_0,q_0,l)
  =
  p_{\theta_a}\!\left(a_{1:H}\mid \mathcal{C}_{\mathrm{base}}\right).
  \label{eq:base}
\end{equation}
During training, the future-observation prediction loss updates the video backbone and shapes the hidden states from which the video KV cache is computed.

\method{} retains the future-observation prediction objective and adds future 2D point-trajectory and visual-feature prediction during training. The corresponding motion and visual-feature streams maintain separate hidden states and serve different downstream roles. At inference, the per-layer K/V projections of the motion-stream tokens form the motion KV cache $\mathcal{C}_m$, which augments the action expert's original conditioning inputs:
\begin{equation}
  p_\theta(a_{1:H}\mid I_0,q_0,l)
  =
  p_{\theta_a}\!\left(
  a_{1:H}
  \mid
  \mathcal{C}_{\mathrm{base}},\mathcal{C}_m
  \right).
  \label{eq:ours}
\end{equation}
No visual-feature KV cache is included in the action expert's conditioning inputs. The visual-feature prediction loss instead trains the visual-feature stream and the shared video backbone.

\subsection{Dual-Stream Dynamic Branch}

Future video latents, 2D point trajectories, and visual features differ in target space and tensor structure. Target-specific prediction heads separate only their output mappings, whereas assigning a full Transformer expert to each target would duplicate the entire processing path. \method{} instead copies the final $M$ blocks of the video backbone to form a dynamic branch (Fig.~\ref{fig:architecture}), in which the motion and visual-feature streams receive target-specific processing through stream-specific FFNs. The original video path retains future-observation prediction. Section~\ref{sec:structured-mask} defines the attention relations among the token groups.

\subsubsection{Tail Attachment and Branch Input}
Let the video backbone contain $N$ blocks, with the first $N-M$ shared by the original video path and the dynamic branch. The branch consists of equal-width copies of the final $M$ Video DiT blocks, initialized from the corresponding video blocks and updated independently. It predicts future 2D point trajectories and visual features from the reference-frame hidden states produced by block $N-M$, without receiving noised future video tokens. This confines the additional target-specific processing to the copied tail.

\subsubsection{Two Reference Copies}
\method{} creates two copies of the reference-frame latent tokens to distribute the demands of action conditioning and the two additional prediction objectives. The copies are content-identical before positional encoding: action queries attend to one, while the other provides the reference-frame hidden states used by the motion and visual-feature streams. Future video tokens attend to both copies, allowing the future-observation loss to update the video backbone through both. During training, the dynamic branch therefore predicts 2D point trajectories and future visual features from current-observation hidden states already shaped by future-observation supervision.

At the branch entrance, the reference-frame hidden states associated with the second copy are duplicated to initialize both streams from identical values. Different learnable stream embeddings are then added, allowing the copied Video DiT blocks to distinguish the two predictive roles.

\subsubsection{Stream-Specific Feed-Forward Networks}
Sharing one feed-forward network would require the same nonlinear transformation parameters to serve both target spaces. In every copied block, \method{} instead duplicates the original feed-forward network into motion and visual-feature FFNs. The two FFNs start from the same parameters but are updated independently, and each stream always uses its corresponding FFN. This fixed assignment follows modality-specific expert designs~\cite{bao2022vlmo} and provides separate FFN parameters without a learned router, Top-$k$ selection, or a load-balancing objective. The two streams therefore apply distinct nonlinear transformations throughout the copied tail rather than differing only at their prediction heads.

\subsection{Motion and Visual-Feature Supervision}

The motion and visual-feature objectives provide complementary supervision for future dynamics and visual structure. Both losses update the dynamic branch and preceding Video DiT blocks.

\subsubsection{Motion Target}
A frozen CoTracker3 tracker~\cite{karaev2024cotracker3} tracks $G_m$ points placed at uniform grid-cell centers in the reference frame $f_0$ of $I_0$, across $P$ representative future frames spanning the action horizon and including its endpoint. The cumulative displacement target is
\begin{equation}
  \Phi_{bcki}=\frac{\tau_{bci}(f_k)-\tau_{bci}(f_0)}{\gamma}\in\mathbb{R}^{2},
  \qquad k=1,\ldots,P,
  \label{eq:motion-target}
\end{equation}
where $i=1,\ldots,G_m$, $\tau_{bci}(f_k)$ denotes pixel coordinates, and $\gamma$ is the spatial compression factor of the variational autoencoder (VAE). Indices $b$ and $c$ identify the $B$ samples and $C$ camera views. Binary masks $m_{bk}$ and $v_{bcki}$ exclude padded frames and invisible points, respectively. With $w_{bcki}=m_{bk}v_{bcki}$, the prediction loss is normalized within each sample--camera pair:
\begin{equation}
  \mathcal{L}_{\mathrm{motion}}=
  \frac{1}{BC}\sum_{b,c}
  \frac{\sum_{k,i}w_{bcki}\,\frac{1}{2}\lVert\hat\Phi_{bcki}-\Phi_{bcki}\rVert_2^2}
       {\max\!\left(1,\sum_{k,i}w_{bcki}\right)}.
  \label{eq:motion-loss}
\end{equation}
All displacements are measured from $f_0$; visible static points remain valid targets with near-zero displacement.

\subsubsection{Visual-Feature Target}
A frozen DINOv2 ViT-B/14 encoder~\cite{oquab2024dinov2} extracts $G_s$ patch targets $\psi_{bckj}\in\mathbb{R}^{d_\psi}$ per future frame and camera, where $j$ indexes patches. Only the $P$ future frames are used, excluding $f_0$. These targets complement image-space motion supervision with \mbox{object-,} \mbox{part-,} and region-level visual structure. Regressing them encourages the video backbone to retain this structure under low-level appearance variation. Predictions $\hat\psi_{bckj}$ are optimized with a cosine loss using the same frame-validity mask and per-pair normalization:
\begin{equation}
  \mathcal{L}_{\mathrm{feat}}=
  \frac{1}{BC}\sum_{b,c}
  \frac{\sum_{k,j}m_{bk}\bigl(1-\cos(\hat\psi_{bckj},\psi_{bckj})\bigr)}
       {\max\!\left(1,\sum_{k,j}m_{bk}\right)}.
  \label{eq:feature-loss}
\end{equation}
\subsubsection{Prediction Heads}
Each stream uses a separately parameterized mask-token decoder~\cite{zhang2025dreamvla,li2026fomovla}. Each decoder linearly projects its stream's final hidden states and concatenates them with mask tokens carrying fixed 2D sine--cosine positional encodings and learned future-frame embeddings. A shallow Transformer followed by a linear layer maps the decoded mask tokens to a $P\times G_m\times2$ displacement tensor or a $P\times G_s\times d_\psi$ feature tensor. Both heads are omitted at inference.

\subsection{Structured Attention Mask}
\label{sec:structured-mask}

\begin{figure}[t]
  \centering
  \makebox[\linewidth][c]{%
    \includegraphics[width=0.98\linewidth]{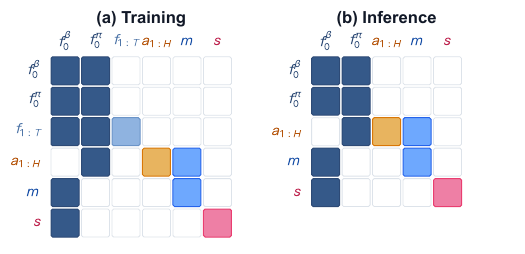}%
  }
  \caption{\textbf{Structured attention masks used by MT-WAM.} Rows are queries and columns are keys; colored cells denote allowed attention. $f_0^\pi$ and $f_0^\beta$ are the reference copies available to action queries and the two branch streams; $f_{1:T}$, $a_{1:H}$, $m$, and $s$ denote the future video, action, motion, and visual-feature groups. Inference omits $f_{1:T}$.}
  \label{fig:masks}
\end{figure}

The structured attention mask extends the video--action mask of the direct-policy interface to the two reference copies and the dynamic branch (Fig.~\ref{fig:masks})~\cite{yuan2026fastwam}. The two reference copies attend to each other but not to noised future video inputs. Future-video attention is restricted to the future sequence and both reference copies. Each branch stream attends within its own sequence and to the reference copy available to both streams, preventing direct mixing of their target-specific hidden states.

Within the action sequence, attention remains bidirectional. Action queries also attend to their reference copy and motion-stream tokens, incorporating future-dynamics information about object motion, interaction progress, and state transitions into the original conditioning context. The motion stream is therefore trained by both the motion and action losses, linking future-dynamics learning to the requirements of action generation.

The visual-feature loss instead updates the visual-feature stream and propagates through its reference-frame inputs to the shared video backbone. Attention between the two reference copies allows this supervision to update backbone parameters that also produce the reference-frame hidden states used for action conditioning. Through this path, future visual-feature prediction supervises the backbone computations that produce the reference-frame keys and values read by the action expert, linking future scene modeling to the visual conditions for action generation (Fig.~\ref{fig:architecture}(c) and (d)).

\subsection{Training Objective}

The full objective is
\begin{equation}
\mathcal{L}=\lambda_{\mathrm{vid}}\mathcal{L}_{\mathrm{vid}}
+\lambda_{\mathrm{act}}\mathcal{L}_{\mathrm{act}}
+\lambda_{\mathrm{motion}}\mathcal{L}_{\mathrm{motion}}
+\lambda_{\mathrm{feat}}\mathcal{L}_{\mathrm{feat}}.
\label{eq:total-loss}
\end{equation}
The first two terms are the original video and action flow-matching objectives~\cite{lipman2023flow,liu2022rectified,yuan2026fastwam}. They act on future video latents and action chunks, respectively, with independently sampled flow times. The motion and visual-feature prediction-head outputs are compared with targets extracted from future observations. Unlike the flow-matching terms, these additional losses use neither noisy interpolation nor flow-time weighting. The video backbone, action expert, and dynamic branch are optimized jointly, while the pretrained VAE and text encoder remain frozen. Appendix~\ref{app:training} reports the loss weights.

\subsection{Inference}

At inference (Fig.~\ref{fig:architecture}(d)), \method{} omits the motion and visual-feature prediction heads and skips future video prediction. The video backbone processes both reference copies in a single pass, after which the dynamic branch computes the motion-stream hidden states from the reference-frame hidden states available to both branch streams. These computations populate the per-layer video and motion KV caches before action denoising. At every denoising step, the action expert reuses both caches according to the structured attention mask. Because the training mask already prevents action queries from attending to noised future video tokens, removing these tokens at inference does not change the information available for action generation (Fig.~\ref{fig:masks}). The video backbone and dynamic branch are each evaluated once per replan, and \method{} directly generates actions without explicitly generating future observations.

\section{Experiments}

\begin{figure*}[t]
  \centering
  \includegraphics[width=\textwidth]{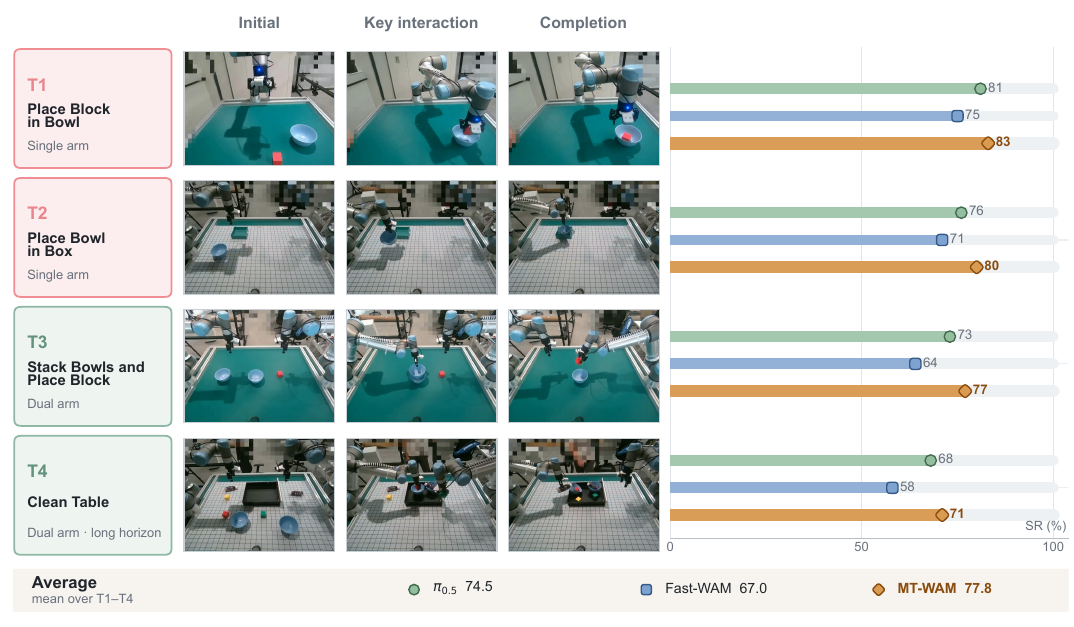}
  \caption{\textbf{Real-world tasks and success rates (\%).} Representative initial, key-interaction, and completion frames for T1--T4 are aligned with task-level SRs. All methods use the same 400 demonstrations and 10-epoch training budget, and are evaluated over the same initial configurations with 100 trials per task; Average is the arithmetic mean over T1--T4.}
  \label{fig:realworld-tasks-results}
\end{figure*}

\subsection{Experiment Setup}
\label{sec:experiment-setup}

We evaluated \method{} on LIBERO, \liberoplus{}, and \robotwin{} Clean2Rand (C2R), as well as on four real-world single- and dual-arm manipulation tasks, using task success rate (SR) as the primary metric. In Tables~\ref{tab:libero-full}--\ref{tab:c2r-full}, PT denotes additional embodied policy pretraining beyond benchmark-specific training. Model initialization, training configurations, and training cost are detailed in Appendix~\ref{app:training}.

\subsubsection{LIBERO and LIBERO-Plus}
We evaluated all methods on LIBERO~\cite{liu2023libero} and \liberoplus{}~\cite{fei2025liberoplus} following their standard protocols. Benchmark-specific training used LIBERO demonstrations without augmented \liberoplus{} data. We report the four-suite Average on LIBERO and the Overall SR across all 10,030 \liberoplus{} variants.

\subsubsection{RoboTwin 2.0}
\robotwin{} is a large-scale simulated benchmark for bimanual robot manipulation~\cite{chen2025robotwin2}. We used all 50 Aloha AgileX tasks, which require coordinated dual-arm control across diverse object layouts and scene conditions. In the multi-task setting, we trained one policy per method for 5 epochs on demonstrations from all tasks, using both Clean and Rand demonstrations. We evaluated each method under both conditions over 100 trials per task and report the average SR across the 50 tasks for each condition, together with their arithmetic mean.

\subsubsection{RoboTwin 2.0 Clean2Rand}
C2R evaluates generalization to Random environments without Random action supervision~\cite{chen2026mvwam}. VLA methods follow a single-stage protocol and are fine-tuned on 50 Clean paired demonstrations per task. WAMs follow a two-stage protocol. Stage~1 pretrains each WAM's video component for 5 epochs on video sequences from 50 Clean and 500 Random demonstrations per task; for \method{}, this stage also trains the dynamic branch. The resulting weights initialize the complete policy, which is then jointly fine-tuned for 5 epochs using only 50 Clean paired demonstrations per task. Random demonstrations therefore provide video-based predictive supervision but no action supervision. Each method is evaluated for 100 Clean and 100 Random rollouts on each of the 50 tasks, totaling 10,000 rollouts per method. We report Clean and Random SR, the absolute Clean-to-Random drop in percentage points, and the descriptive Random retention $100\,\mathrm{SR}_{\mathrm{Random}}/\mathrm{SR}_{\mathrm{Clean}}$.

\begingroup
\begin{table}[!b]
\caption{Success rates (\%) on LIBERO}
\label{tab:libero-full}
\centering
\footnotesize
\setlength{\tabcolsep}{2.5pt}
\tablehighlightsetup
\begin{tabular}{@{}lc*{4}{r}>{\columncolor{tablehighlight}}r@{}}
\toprule
Method & PT & Spatial & Object & Goal & Long & Average \\
\midrule
OpenVLA~\cite{kim2024openvla} & \cmark & 84.7 & 88.4 & 79.2 & 53.7 & 76.5 \\
$\pi_0$~\cite{black2024pi0} & \cmark & 96.8 & 98.8 & 95.8 & 85.2 & 94.1 \\
$\pi_{0.5}$~\cite{black2025pi05} & \cmark & \second{98.8} & 98.2 & \second{98.0} & 92.4 & 96.9 \\
LingBot-VA~\cite{li2026lingbotva} & \cmark & 98.5 & 99.6 & 97.2 & \best{98.5} & \best{98.5} \\
Motus~\cite{bi2025motus} & \cmark & 96.8 & \second{99.8} & 96.6 & \second{97.6} & 97.7 \\
\fastwam{}~\cite{yuan2026fastwam} & \xmark & 98.2 & \best{100.0} & 97.0 & 95.2 & 97.6 \\
\method{} & \xmark & \best{99.3} & 99.2 & \best{98.3} & 96.0 & \second{98.2} \\
\bottomrule
\end{tabular}
\end{table}
\endgroup

\begin{table*}[!t]
\caption{Success rates (\%) on \liberoplus{}}
\label{tab:liberoplus-full}
\centering
\footnotesize
\setlength{\tabcolsep}{3.5pt}
\renewcommand{\arraystretch}{1.08}
\tablehighlightsetup
\begin{tabular*}{\textwidth}{@{\extracolsep{\fill}}lc*{7}{r}>{\columncolor{tablehighlight}}r@{}}
\toprule
Method & PT & Cam. & Rob. & Lang. & Light & Bg. & Noise & Lay. & Ovr. \\
\midrule
UniVLA~\cite{bu2025univla} & \cmark & 1.80 & 46.20 & 69.60 & 69.00 & 81.00 & 21.20 & 31.90 & 42.87 \\
OpenVLA-OFT~\cite{kim2025openvlaoft} & \cmark & 56.40 & 31.90 & 79.50 & 88.70 & \best{93.30} & 75.80 & 74.20 & 69.59 \\
$\pi_0$~\cite{black2024pi0} & \cmark & 13.80 & 6.00 & 58.80 & 85.00 & 81.40 & 79.00 & 68.90 & 53.63 \\
$\pi_0$-FAST~\cite{pertsch2025fast} & \cmark & \best{65.10} & 21.60 & 61.00 & 73.20 & 73.20 & 74.40 & 68.80 & 61.59 \\
WorldVLA~\cite{cen2025worldvla} & \cmark & 0.10 & 27.90 & 41.60 & 43.70 & 17.10 & 10.90 & 38.00 & 25.03 \\
\addlinespace[1pt]
\fastwam{}~\cite{yuan2026fastwam} & \xmark & 16.26 & 44.13 & 66.82 & 79.77 & 52.60 & 38.54 & 61.38 & 49.86 \\
Fast-WAM-Joint~\cite{yuan2026fastwam} & \xmark & 39.59 & 60.90 & \second{92.32} & \best{94.57} & 57.62 & 58.59 & \second{80.52} & 68.41 \\
SG-WAM~\cite{zhao2026sgwam} & \xmark & \second{58.60} & 48.90 & 81.40 & 89.80 & \second{86.10} & \best{80.70} & 74.20 & \second{73.00} \\
ST-WAM~\cite{wang2026stwam} & \xmark & 55.40 & 60.10 & 79.30 & 93.00 & 74.20 & \second{79.50} & 74.30 & 72.80 \\
4D-WAM~\cite{yang2026four_d_wam} & \xmark & 45.15 & \best{64.26} & 90.63 & \second{94.29} & 57.71 & 69.08 & 79.21 & 71.01 \\
\method{} & \xmark & 52.70 & \second{61.10} & \best{94.70} & 93.60 & 64.80 & 71.30 & \best{81.00} & \best{73.66} \\
\bottomrule
\end{tabular*}
\end{table*}

\subsubsection{Real-World Experiments}
We evaluated four tasks using UR5 manipulators in single-arm and dual-arm setups, with all visual observations captured by Intel RealSense D405 cameras (Fig.~\ref{fig:realworld-tasks-results}). Place Block in Bowl (T1) and Place Bowl in Box (T2) use one arm; Stack Bowls and Place Block (T3) and the long-horizon Clean Table task (T4) use two arms. Each task contained 100 demonstrations, evenly divided between solid and patterned tabletop backgrounds. We pooled all 400 demonstrations across tasks, camera viewpoints, and background types to train one multi-task policy per method for 10 epochs. Each method was evaluated over 100 trials per task, with a solid or patterned tabletop background selected randomly for every trial and the same initial configurations used across methods. We report task-level SR and the arithmetic mean across the four tasks. Appendix~\ref{app:realworld} provides additional details.

\subsection{Main Results}

\subsubsection{LIBERO and LIBERO-Plus}

On LIBERO, \method{} achieved an average SR of 98.2\%, compared with 97.6\% for \fastwam{} (+0.6 percentage points; Table~\ref{tab:libero-full}). It improved LIBERO-Spatial, LIBERO-Goal, and LIBERO-Long by 1.1, 1.3, and 0.8 percentage points, respectively, while scoring 0.8 points lower on LIBERO-Object. These small suite-level differences indicate comparable performance on the standard LIBERO tasks.

On \liberoplus{}, \method{} achieved an Overall SR of 73.66\% without additional embodied policy pretraining, the highest among the methods compared in Table~\ref{tab:liberoplus-full} and 23.80 percentage points above \fastwam{}. The gains extended across all seven perturbation axes, with the largest improvements under Camera, Noise, Language, and Layout, where SR increased by 36.44, 32.76, 27.88, and 19.62 percentage points, respectively. These results, together with the comparable performance on standard LIBERO, show that \method{} maintains performance on the original tasks and improves robustness when observations, instructions, or scene conditions change. These findings support the central premise of \method{}: improving the one-pass predictive representation along future-dynamics and visual-structure dimensions can strengthen action generation without explicitly predicting future video at inference.

\subsubsection{RoboTwin 2.0 and Clean2Rand}

In the \robotwin{} Clean/Rand setting, \method{} achieved an average SR of 93.22\% without additional embodied policy pretraining, exceeding \fastwam{} by 1.39 percentage points (Table~\ref{tab:robotwin-full}). It achieved the highest SR among the compared methods in both conditions, with similar performance on Clean and Rand, showing consistent control across conditions included in action-supervised training.

\robotwin{} Clean2Rand instead evaluates generalization to Random environments without Random action supervision. \method{} achieved a Random SR of 19.40\%, exceeding \fastwam{} by 13.10 percentage points (Table~\ref{tab:c2r-full}). HALO achieved a higher Clean SR, while HALO and BagelVLA achieved higher Random SRs; both methods use additional embodied policy pretraining. The larger advantage over \fastwam{} on Random than on Clean supports the central premise of \method{}: improving the one-pass predictive representation through complementary future supervision and target-specific processing can strengthen action generation beyond Clean action supervision.

\begingroup
\begin{table}[!t]
\caption{Success rates (\%) on \robotwin{} Clean/Rand}
\label{tab:robotwin-full}
\centering
\footnotesize
\setlength{\tabcolsep}{3.2pt}
\tablehighlightsetup
\begin{tabular}{@{}lc*{2}{r}>{\columncolor{tablehighlight}}r@{}}
\toprule
Method & PT & Clean & Rand & Average \\
\midrule
$\pi_0$~\cite{black2024pi0} & \cmark & 65.92 & 58.40 & 62.16 \\
$\pi_{0.5}$~\cite{black2025pi05} & \cmark & 82.74 & 76.76 & 79.75 \\
ABot-M0~\cite{yang2026abotm0} & \xmark & 81.20 & 80.40 & 80.80 \\
Motus~\cite{bi2025motus} & \cmark & 88.66 & 87.02 & 87.84 \\
Motus from Wan 2.2~\cite{bi2025motus} & \xmark & 77.56 & 77.00 & 77.28 \\
LingBot-VA~\cite{li2026lingbotva} & \cmark & \second{92.90} & 91.50 & \second{92.20} \\
\fastwam{}~\cite{yuan2026fastwam} & \xmark & 91.88 & \second{91.78} & 91.83 \\
\method{} & \xmark & \best{93.18} & \best{93.26} & \best{93.22} \\
\bottomrule
\end{tabular}
\end{table}
\endgroup

\begingroup
\begin{table}[!t]
\caption{Success rates (\%) on \robotwin{} Clean2Rand}
\label{tab:c2r-full}
\centering
\footnotesize
\setlength{\tabcolsep}{1.8pt}
\tablehighlightsetup
\setlength{\extrarowheight}{0.1pt}
\begin{tabular}{@{}lcr>{\columncolor{tablehighlight}}rrr@{}}
\toprule
Method & PT & Clean & Random & \shortstack{Absolute\\drop (pp)} & \shortstack{Random\\retention (\%)} \\
\midrule
$\pi_0$~\cite{black2024pi0} & \cmark & 46.40 & 16.30 & 30.10 & 35.1 \\
RDT~\cite{liu2024rdt} & \cmark & 34.50 & 13.70 & 20.80 & 39.7 \\
UP-VLA~\cite{zhang2025upvla} & \cmark & 52.90 & 15.20 & 37.70 & 28.7 \\
Diffusion Policy (DP)~\cite{chi2023diffusionpolicy} & \xmark & 28.00 & 0.60 & 27.40 & 2.1 \\
HALO~\cite{shou2026halo} & \cmark & \best{80.50} & \best{26.40} & 54.10 & 32.8 \\
BagelVLA~\cite{hu2026bagelvla} & \cmark & 75.30 & \second{20.50} & 54.80 & 27.2 \\
\fastwam{}~\cite{yuan2026fastwam} & \xmark & 71.90 & 6.30 & 65.60 & 8.8 \\
StarVLA~\cite{ye2026starvla} & \xmark & 65.72 & 11.68 & 54.04 & 17.8 \\
\method{} & \xmark & \second{75.56} & 19.40 & 56.16 & 25.7 \\
\bottomrule
\end{tabular}
\end{table}
\endgroup

\subsubsection{Real-World Experiments}
\label{sec:realworld-results}

Fig.~\ref{fig:realworld-tasks-results} aligns representative task progressions with their task-level SRs. Without additional embodied policy pretraining, \method{} achieved an average SR of 77.8\%, compared with 74.5\% for $\pi_{0.5}$, which uses such pretraining, and 67.0\% for \fastwam{}~\cite{yuan2026fastwam}. Among the compared methods, \method{} achieved the highest SR on all four tasks. These tasks span single-arm object placement, multi-step bimanual manipulation, and long-horizon multi-object manipulation. Relative to \fastwam{}, \method{} improved SR by 8, 9, 13, and 13 percentage points on T1 (Place Block in Bowl), T2 (Place Bowl in Box), T3 (Stack Bowls and Place Block), and T4 (Clean Table), respectively. The gains were largest on T3 and T4, which place greater demands on maintaining interaction progress and state-transition information across successive action chunks. Each trial used a randomly selected solid or patterned tabletop background, and the gains\nopagebreak[4] persisted across this background variation. Both \fastwam{}\nopagebreak[4] and \method{} skip explicit future generation at inference, so the improvement does not depend on an additional future video rollout. These results show that complementary future-motion and visual-feature supervision strengthens real-world control while preserving direct action generation.

\subsubsection{Inference Efficiency}

Table~\ref{tab:inference-efficiency} compares inference efficiency on \liberoplus{}. All models were measured under the same inference configuration on a single NVIDIA RTX PRO 6000 96~GB GPU. We report the number of floating-point operations in trillions (TFLOPs).

\begingroup
\begin{table}[!htbp]
\caption{Inference efficiency on \liberoplus{} using one NVIDIA RTX PRO 6000 96~GB GPU}
\label{tab:inference-efficiency}
\centering
\footnotesize
\setlength{\tabcolsep}{2.0pt}
\begin{tabular}{@{}lcrrr@{}}
\toprule
Method & \shortstack{Future\\video} & \shortstack{Overall\\SR (\%)} & \shortstack{Latency\\(ms)} & TFLOPs \\
\midrule
\fastwam{}~\cite{yuan2026fastwam} & \xmark & 49.86 & \textbf{296.325} & \textbf{3.736949} \\
Fast-WAM-Joint~\cite{yuan2026fastwam} & \cmark & 68.41 & 503.729 & 30.672534 \\
\method{} & \xmark & \textbf{73.66} & 313.480 & 5.142984 \\
\bottomrule
\end{tabular}
\end{table}
\endgroup

\method{} achieves the highest Overall SR among the three methods in Table~\ref{tab:inference-efficiency}. Compared with \fastwam{}, it improves Overall SR by 23.80 percentage points, with increases of 5.79\% in latency and 37.63\% in FLOPs. Thus, the additional computation is accompanied by a substantial improvement in control performance, while inference latency remains close to \fastwam{}.

Compared with Fast-WAM-Joint, \method{} uses approximately one-sixth of the floating-point operations while achieving an Overall SR 5.25 percentage points higher. Our design improves the one-pass predictive representation, enabling \method{} to achieve a higher Overall SR than Fast-WAM-Joint without restoring full future-video denoising at inference.

\subsection{Ablation Study}
\label{sec:ablations}

\begin{figure}[!htbp]
  \centering
  \includegraphics[width=\columnwidth]{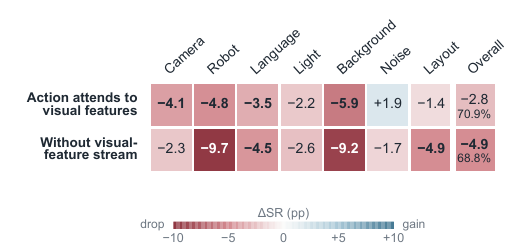}
  \caption{\liberoplus{} ablation effects relative to \method{}. Each cell reports the change in SR in percentage points; the Overall SR entries additionally report the absolute SR. \method{} achieves an Overall SR of 73.7\%.}
  \label{fig:liberoplus-ablation}
\end{figure}

Our ablations address three questions about the design of \method{}: whether action queries should attend to the visual-feature-stream tokens, whether the visual-feature stream contributes to control performance, and how performance varies with the branch depth $M$. Fig.~\ref{fig:liberoplus-ablation} summarizes the first two comparisons, and Fig.~\ref{fig:branch-depth} reports the branch-depth study. Each comparison changes the design factor specified by its corresponding question and follows the \liberoplus{} protocol in Section~\ref{sec:experiment-setup}.

\subsubsection{Should Action Queries Attend to the Visual-Feature Stream}
The controlled comparison changes only one attention rule: action queries attend to the visual-feature-stream tokens as well as the motion-stream tokens, while the two streams, prediction objectives, and parameter count remain unchanged. The Overall SR falls from 73.7\% to 70.9\%, a reduction of 2.8 percentage points, and the SR decreases on six of the seven perturbation axes. The largest drops occur under Background (\(-5.9\)), Robot (\(-4.8\)), and Camera (\(-4.1\)). Noise is the only exception, with a 1.9-point gain. Because no other architectural component changes, the comparison isolates the effect of adding visual-feature-stream tokens to the action-conditioning context. The result shows that this additional visual-feature conditioning does not improve action generation and supports excluding it.

\subsubsection{Does the Visual-Feature Stream Contribute}
Removing the visual-feature stream lowers the Overall SR from 73.7\% to 68.8\%, a reduction of 4.9 percentage points, and decreases the SR on all seven perturbation axes. The largest reductions occur under Robot (\(-9.7\)), Background (\(-9.2\)), and Layout (\(-4.9\)). The reductions span changes in robot initial states, object layouts, language instructions, lighting, backgrounds, and image noise. The visual-feature stream therefore contributes across both structural and visual perturbations.

Taken together, the two ablations support the intended asymmetric roles of the branch streams. The visual-feature stream improves control through its training objective and target-specific processing, but allowing action queries to attend to its tokens reduces the Overall SR. The visual-feature stream therefore refines the video backbone during training, while the motion stream supplies the additional future-dynamics information used directly for action generation.

\begin{figure}[!t]
  \centering
  \includegraphics[width=\columnwidth]{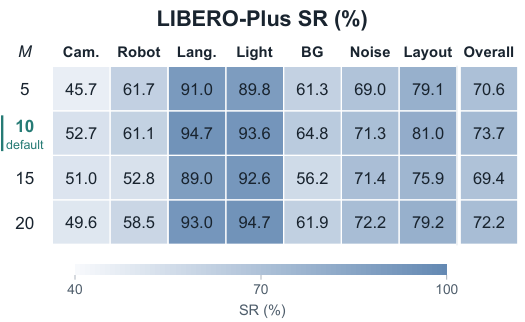}
  \caption{Success rates (\%) across branch depths on \liberoplus{}.}
  \label{fig:branch-depth}
\end{figure}

\subsubsection{How Does Performance Vary with Branch Depth?}
We vary $M$ while keeping all other settings fixed (Fig.~\ref{fig:branch-depth}).

Increasing $M$ starts the dynamic branch at an earlier Video DiT block. It allocates more copied blocks and parameters to target-specific processing, but leaves fewer video blocks shared before the branch. Among the evaluated depths, $M=10$ achieves the highest Overall SR at 73.7\%, although no single configuration leads every perturbation axis. The axis-specific leaders are distributed across the settings: $M=5$ performs best under Robot and $M=20$ under Light and Noise, whereas $M=10$ leads Camera, Language, Background, and Layout. The Overall SR advantage of $M=10$ therefore reflects a balanced profile rather than a peak on one axis. It also uses half as many copied blocks as $M=20$ while achieving an Overall SR 1.5 percentage points higher. Performance is thus non-monotonic with branch depth. We use $M=10$ as the default because it attains the highest Overall SR among the evaluated depths with half as many copied blocks as $M=20$.

\FloatBarrier

\section{Discussion and Conclusion}

We introduced \method{}, which improves the one-pass predictive representation along future-dynamics and visual-structure dimensions through a lightweight dynamic branch. Gains over \fastwam{}~\cite{yuan2026fastwam} are more pronounced under distribution shift and in real-world manipulation: 23.80 percentage points on \liberoplus{}, 13.10 points in the \robotwin{} Clean2Rand Random condition, and 10.8 points on average across real-world tasks. The largest real-world gains occur on tasks requiring multi-step object interactions, consistent with the importance of interaction progress and state-transition information.

The ablations support the asymmetric roles of the two streams. The visual-feature pathway contributes through training the backbone's scene encoding, while motion-stream tokens provide the additional dynamics conditions used directly for action generation. Removing the visual-feature stream lowers \liberoplus{} Overall SR by 4.9 percentage points, whereas allowing action queries to attend to its tokens lowers it by 2.8 points. These results indicate that a predictive stream can improve control through its contribution to training without its tokens serving as additional action conditions.

The dynamic branch does more than add capacity: its starting point sets the boundary between shared video representation learning and target-specific processing. The non-monotonic depth results suggest that control benefits from balancing shared predictive features with independent transformations for heterogeneous targets. A copied tail provides these target-specific transformations without duplicating the full video-processing hierarchy for each target, limiting the additional computation required for specialization. The efficiency results support the practicality of this allocation: \method{} strengthens action generation with a modest latency increase over \fastwam{} and lower computation than joint video--action denoising.

\textit{Limitations.} The 2D point-trajectory target describes image-space motion but does not directly encode metric 3D geometry or contact states. Adding these targets could extend the dynamic branch's predictive information. Evaluation covers one pretrained video backbone and UR5 single- and dual-arm configurations, without additional embodied policy pretraining. Further evaluation across backbones, embodiments, and longer-horizon tasks would examine the scope of the design, while large-scale embodied policy pretraining would test its complementarity with data scaling.

Overall, \method{} shows that, for WAMs that omit explicit future generation, translating predictive supervision into control gains depends on making the one-pass predictive representation better serve action generation. Results from simulation, real-world experiments, controlled ablations, and efficiency analysis jointly show that complementary supervision for future dynamics and visual structure, together with the asymmetric design, strengthens direct action generation without restoring iterative future-video denoising.

\appendices

\section{Training Details}
\label{app:training}

The video backbone is initialized from the Wan 2.2-TI2V-5B checkpoint~\cite{wan2025}.\footnote{\url{https://huggingface.co/Wan-AI/Wan2.2-TI2V-5B}} The action expert uses the Action DiT architecture and is also initialized from the Wan 2.2 Video DiT. Exact-shape backbone tensors are copied directly, while size-mismatched tensors are linearly interpolated with width-dependent scaling. The action encoder and output head are randomly initialized.

All models are trained on 8 NVIDIA A100 GPUs with DeepSpeed ZeRO-1 and bfloat16 precision. We use AdamW with $\beta_1=0.9$, $\beta_2=0.95$, learning rate $1\times10^{-4}$, and weight decay $1\times10^{-2}$. The learning rate follows a warmup-cosine schedule: linear warmup spans the first $0.05T_{\mathrm{total}}$ steps, and the minimum learning rate is $0.01$ times the initial rate. Gradients are clipped to a global norm of $1.0$. The loss weights are $\lambda_{\mathrm{vid}}=\lambda_{\mathrm{act}}=1.0$, $\lambda_{\mathrm{motion}}=0.5$, and $\lambda_{\mathrm{feat}}=0.25$. Table~\ref{tab:training-configs} summarizes the dataset-specific configurations.

The C2R protocol has two stages. Stage~1 uses video sequences from 50 Clean and 500 Random demonstrations per task to pretrain the video backbone and dynamic branch for 5 epochs. Stage~2 starts from this checkpoint and jointly fine-tunes the complete policy for 5 epochs using only 50 Clean paired demonstrations per task. Random demonstrations therefore provide video-based predictive supervision in Stage~1 but no action supervision. Table~\ref{tab:training-cost} reports the training cost and global batch size for each protocol.

\section{Real-World Experiment Details}
\label{app:realworld}

We used UR5 manipulators in single-arm and dual-arm setups, with all visual observations captured by Intel RealSense D405 cameras. Place Block in Bowl and Place Bowl in Box used one arm; Stack Bowls and Place Block and the long-horizon Clean Table task used two arms. We collected 100 demonstrations per task, evenly divided between solid and patterned tabletop backgrounds. The resulting 400 demonstrations spanned multiple camera viewpoints and were used to train one multi-task policy per method for 10 epochs. Training took approximately 17 hours on 8 NVIDIA A100 GPUs with a global batch size of 48. Each method was evaluated over 100 trials per task. Every trial used a randomly selected solid or patterned background, and all methods shared the same initial configurations. We report task-level SR and the arithmetic mean over the four tasks.

\begin{table}[!htbp]
\caption{Dataset-specific training configurations}
\label{tab:training-configs}
\centering
\normalsize
\setlength{\tabcolsep}{3pt}
\renewcommand{\arraystretch}{1.08}
\renewcommand{\tabularxcolumn}[1]{m{#1}}
\begin{tabularx}{\columnwidth}{@{}>{\raggedright\arraybackslash}m{0.30\columnwidth}*{3}{>{\centering\arraybackslash}X}@{}}
\toprule
\multirow{2}{*}{Configuration} & \multirow{2}{*}{LIBERO} & \multicolumn{2}{c}{\robotwin{}} \\
\cmidrule(lr){3-4}
& & Clean/Rand & C2R \\
\midrule
Camera views & 2 & 3 & 3 \\
Camera layout & Horizontal & \shortstack{Wrist-\\horizontal\\+ vertical} & \shortstack{Wrist-\\horizontal\\+ vertical} \\
Input resolution & $224\times448$ & $384\times320$ & $384\times320$ \\
Observation window & 33 & 33 & 33 \\
Action chunk length & 32 & 32 & 32 \\
Action dimension & 7 & 14 & 14 \\
Training epochs & 10 & 5 & \shortstack{Stage 1: 5\\Stage 2: 5} \\
\bottomrule
\end{tabularx}
\end{table}

\begin{table}[!htbp]
\caption{Training cost and global batch size}
\label{tab:training-cost}
\centering
\normalsize
\setlength{\tabcolsep}{8pt}
\renewcommand{\arraystretch}{1.04}
\begin{tabular}{@{}lcc@{}}
\toprule
Protocol & Time & Batch size \\
\midrule
LIBERO & 16 h & 96 \\
\robotwin{} & 4 days & 48 \\
\robotwin{} C2R Stage~1 & 4 days & 48 \\
\robotwin{} C2R Stage~2 & 18 h & 48 \\
Real-world multi-task & 17 h & 48 \\
\bottomrule
\end{tabular}
\end{table}

\FloatBarrier

\bibliographystyle{IEEEtran}
\bibliography{references}

\end{document}